# SIMILAR DATA POINTS IDENTIFICATION WITH LLM: A HUMAN-IN-THE-LOOP STRATEGY USING SUMMARIZATION AND HIDDEN STATE INSIGHTS

Xianlong Zeng[1], Yijing Gao[2], Ang Liu[3], and Fanghao Song[1]

[1]Electrical Engineering and Computer Science, Ohio University, Athens, OH 45701
[2]Data Science Institute, Brown University, Providence, RI 02912
[3]Global Urban Studies Program, Rutgers University, New Brunswick, NJ 08901
xz926813@ohio.edu

## ABSTRACT

*This study introduces a simple yet effective method for identifying similar data points across non-free text domains, such as tabular and image data, using Large Language Models (LLMs). Our two-step approach involves data point summarization and hidden state extraction. Initially, data is condensed via summarization using an LLM, reducing complexity and highlighting essential information in sentences. Subsequently, the summarization sentences are fed through another LLM to extract hidden states, serving as compact, feature-rich representations. This approach leverages the advanced comprehension and generative capabilities of LLMs, offering a scalable and efficient strategy for similarity identification across diverse datasets. We demonstrate the effectiveness of our method in identifying similar data points on multiple datasets. Additionally, our approach enables non-technical domain experts, such as fraud investigators or marketing operators, to quickly identify similar data points tailored to specific scenarios, demonstrating its utility in practical applications. In general, our results open new avenues for leveraging LLMs in data analysis across various domains.*

## 1. INTRODUCTION

In the ever-evolving landscape of data science, the task of identifying similar data points stands as a cornerstone of numerous advanced applications, from refining the precision of search engines and enhancing recommendation systems to streamlining processes of data deduplication and anomaly detection. This critical function underpins the development and efficiency of machine learning models, influencing their performance and applicability across diverse domains such as healthcare, finance, housing, and e-commerce. However, the digital era's exponential growth in data generation has introduced significant challenges. The vast volumes of data, coupled with its increasing complexity and variety, including structured, unstructured, tabular, and image data, demand innovative approaches that transcend traditional data analysis methods. These challenges are further amplified by the necessity for scalable, efficient, and accurate techniques to manage and analyze this data deluge, highlighting the urgent need for novel methodologies that can adeptly handle the intricacies of similarity identification in such a multifaceted data environment.

Traditional methods for identifying similar data points, such as clustering algorithms, nearest neighbor searches, and multidimensional scaling, while foundational, often grapple with the limitations imposed by the scale and diversity of contemporary datasets. These approaches typically require extensive preprocessing and feature engineering efforts, especially when dealing with non-textual data types, which can be both time-consuming and resource-intensive. Moreover, their effectiveness diminishes as the dimensionality and complexity of the data increase, leading to scalability issues and a decline in accuracy. In the context of non-free text domains, such as tabular data and images, the challenge is further compounded. The intrinsic characteristics of these data types necessitate a more nuanced understanding and representation of data to facilitate meaningful similarity identification. As a result, there is a pressing need for innovative solutions that can accommodate the rich diversity of data forms and scales, enabling efficient and effective identification of similar data points without the burdensome requirements of traditional methodologies.

Enter Large Language Models (LLMs), a revolutionary stride in artificial intelligence that offers a promising solution to these challenges. LLMs, with their advanced comprehension and generative capabilities, have demonstrated unprecedented success in understanding and generating human-like text. Their architecture, designed to process vast amounts of information and learn nuanced patterns within, makes them particularly adept at handling a wide range of data types beyond mere text. This adaptability stems from their ability to abstract and contextualize information, allowing for the processing of tabular data, images, and more through natural language-like representations. By leveraging LLMs, we can transcend traditional barriers in data analysis, utilizing their robust, pre-trained models to perform tasks such as summarization, translation, and even sentiment analysis across various data modalities. The potential of LLMs to revolutionize the identification of similar data points lies not just in their scalability and efficiency but also in their capacity to provide deeper insights into the data, paving the way for more sophisticated and nuanced approaches to understanding and leveraging data similarity.

Building on the foundational capabilities of Large Language Models (LLMs), our study introduces a novel, two-step approach designed to harness their power for identifying similar data points across diverse, non-free text domains such as tabular and image data. The first step of our methodology involves the use of an LLM to generate summarizations of data points. This process effectively reduces the complexity of the data, distilling it into its most essential features and information in a natural language format that is inherently more manageable for analysis. Following this, we employ another LLM to process these summarizations, extracting hidden state representations. These representations serve as compact yet rich, feature vectors that encapsulate the essence of the data points. By leveraging these advanced neural network capabilities, our approach not only simplifies the data but also enriches the analysis, enabling a scalable and efficient strategy for identifying similarities. This innovative method underscores the versatility and potential of LLMs in data analysis, offering a groundbreaking solution to the challenges posed by the multifaceted nature of modern datasets. By simplifying the process of identifying similar data points through LLM-driven summarization and hidden state extraction, our approach democratizes access to advanced data analysis techniques. This accessibility empowers professionals across various fields, from fraud investigators in finance to marketing strategists in retail, enabling them to pinpoint relevant data patterns and insights without deep technical expertise in data science or machine learning. In summary, the key contributions of our paper are as follows:

- We introduce a novel two-step approach leveraging LLMs for summarization and hidden state extraction to identify similar data points, effectively bridging the gap between complex datasets and actionable insights.
- By reducing data complexity through summarization before extracting dense, feature-rich representations, our approach offers a scalable and efficient solution for analyzing large datasets. Empowers domain experts without deep technical backgrounds, enabling them to leverage advanced data analysis techniques for informed decision-making.
- In our experiments, we demonstrated the effectiveness of our method, which paves the way for further exploration into the capabilities of LLMs in data analysis and beyond, offering a rich avenue for future innovation.

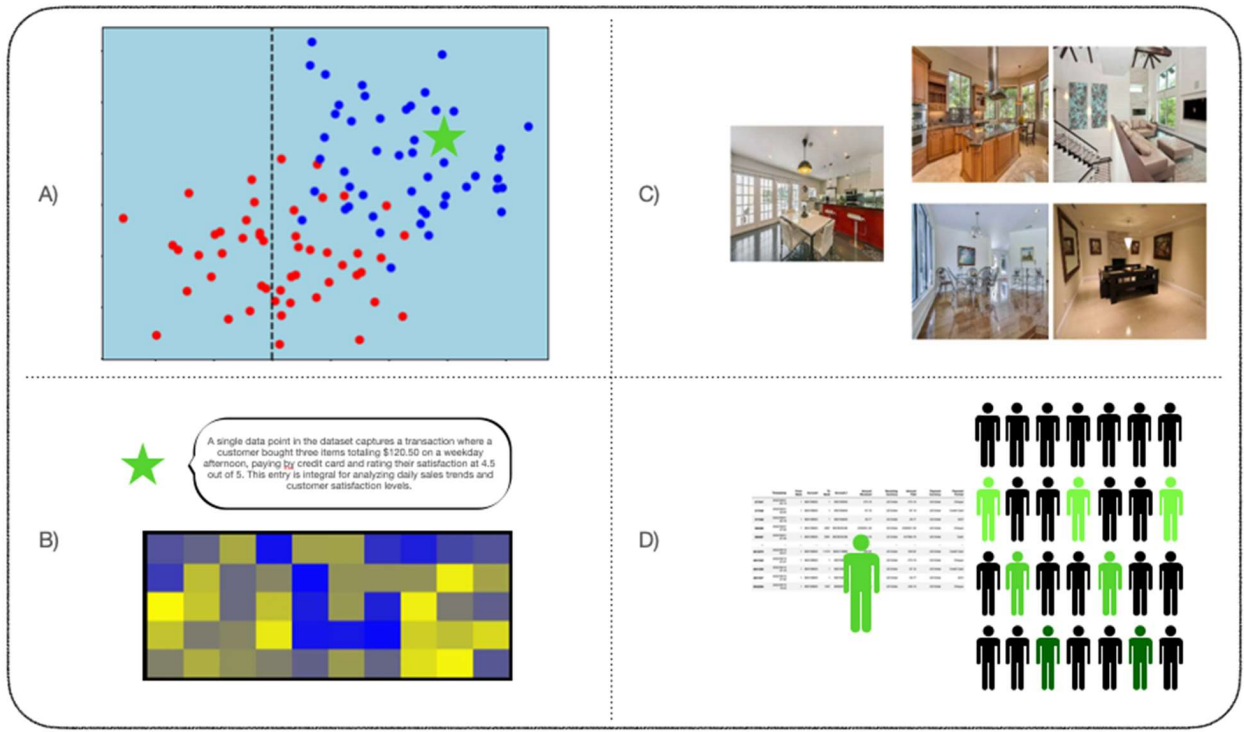

Figure 1. A high-level illustration of our proposed method. A) For a point-of-interest (green star) in the dataset, B) our method first leverages an summarization LLM to extract the data profile based on human's interest, another embedding LLM is then applied to extract the hidden state of the data point, C) example illustration in the image dataset, D) example illustration in the tableau dataset.

The remainder of this paper is structured to systematically explore and substantiate our findings. Section 2 explores related work, offering a review of existing methodologies for identifying similar data points and the role of Large Language Models (LLMs) in data analysis, establishing the foundation upon which our research builds. Section 3 introduces our innovative two-step methodology, detailing the process of data summarization and hidden state extraction using LLMs and elucidating the theoretical underpinnings and practical applications of this approach. Section 4 presents a comprehensive analysis of our methodology's application across various domains, including tabular data and image data, demonstrating the versatility and effectiveness of our approach through empirical evidence. Section 6 discusses the limitations and future work, suggesting directions for future research to further leverage and expand upon the capabilities of LLMs in identifying similar data points and beyond. Finally, Section 5 concludes the paper, summarizing the key insights and contributions of our study.

## 2. RELATED WORK

### 2.1. Deep Learning, Large Language Model, and Prompting

Deep learning, a subset of machine learning based on artificial neural networks, has profoundly impacted various fields, offering insights and advancements that were previously unattainable. LeCun, Bengio, and Hinton [12] were pivotal in highlighting the power of deep learning, providing comprehensive insights into the capabilities of deep neural networks, particularly in image and speech recognition. Their foundational work set the stage for numerous applications across different domains. For instance, Silver et al. [13] showcased the potential of deep reinforcement learning in the realm of board games, specifically Go, where their system, AlphaGo, defeated a world champion. This landmark achievement not only demonstrated the strategic capability of deep learning algorithms but also spurred further research into complex problem-solving. In healthcare, deep learning has been instrumental in advancing diagnostic procedures. Zeng et al. [18-22] demonstrated that deep neural networks show significant performance on claims data for medical outcome prediction. Similarly, Jin et al. [23] and Lin et al. [24] apply deep learning models for clinical outcome prediction. In natural language processing (NLP), Vaswani et al. [14] introduced the Transformer model, a novel neural network architecture eschewing recurrence in favor of attention mechanisms, which has since become the backbone of many state-of-the-art NLP systems. The transformative effects of deep learning continue to permeate through more specialized applications as well, such as autonomous driving, where Bojarski et al. [15] detailed the use of convolutional neural networks for direct perception in self-driving cars.

The efficacy of LLMs is often contingent on the quality of the prompting mechanisms used to guide their output. In their influential paper, Brown et al. [1] unveiled GPT-3, demonstrating the model's ability to perform a variety of tasks without specific tuning by leveraging effective prompting techniques. This notion of "prompt engineering" or "prompt design" has become a critical area of research, with subsequent studies examining how prompts can be optimized to improve the performance of LLMs across tasks [2]. Further exploration into the interaction between LLMs and prompting has revealed that the architecture of these models inherently encodes a vast range of knowledge, accessible through the right prompts. Recent works by Shin et al. [3] have presented methodologies for zero-shot and few-shot learning, where LLMs perform tasks without explicit prior training, purely based on the context provided in a prompt. This area of prompting as a means of accessing the "in-context learning" capabilities of LLMs continues to expand, with researchers investigating the limits and potential of these interactions [6]. The application of prompting in LLMs has also extended to more complex tasks, such as code generation and data analysis. Studies by Chen et al.[4] on Codex have showcased how prompts can guide LLMs to generate code from natural language descriptions, further underscoring the versatility of prompting. In the realm of data analysis, where structured and unstructured data converge, the role of prompting becomes even more nuanced. Researchers are beginning to understand how LLMs can be prompted to interpret and generate insights from data, revealing patterns and anomalies [5]. The body of work on LLMs and prompting illustrates a dynamic field where the frontier is constantly advancing. As researchers continue to unveil the intricacies of these models, the art of prompting stands as a testament to the creative synergy between human ingenuity and artificial intelligence [23].

### 2.2. Human-in-the-loop Data Analysis

Human-in-the-loop data analysis has emerged as a critical field, balancing the scale and efficiency of algorithmic processing with the nuanced understanding of human analysts [11]. Holzinger et al. posited the need for integrative machine learning approaches that involve humans [7], especially in the medical informatics field, where interpretability is as crucial as predictive accuracy. Following this, researchers have explored HITL methodologies in various domains, finding that human expertise not only improves the performance of machine learning models but also ensures their decisions are more transparent and justifiable. For instance, Amershi et al. [8] discussed how interactive machine learning systems benefit from iterative input from users, enhancing the overall learning process. Recent advances by Kim et al. [9] in explainable AI have demonstrated how HITL can lead to the development of systems that provide explanations aligned with human reasoning, fostering trust and enabling users to effectively manage automated systems. This synergy between human and machine intelligence is pivotal for applications ranging from data categorization to complex decision-making processes, where Cai et al. [10] illustrated that HITL approaches could significantly reduce the time and expertise required to create and maintain machine learning models. Collectively, these works underscore the evolving role of HITL in data analysis, where the collaboration between human intuition and algorithmic power is optimized to improve outcomes across various analytical tasks.

## 3. METHOD

This section delineates the methodology employed to identify similar data points across various non-free text domains, leveraging the capabilities of Large Language Models (LLMs). Our approach comprises two primary stages: data summarization and hidden state extraction, followed by similarity analysis.

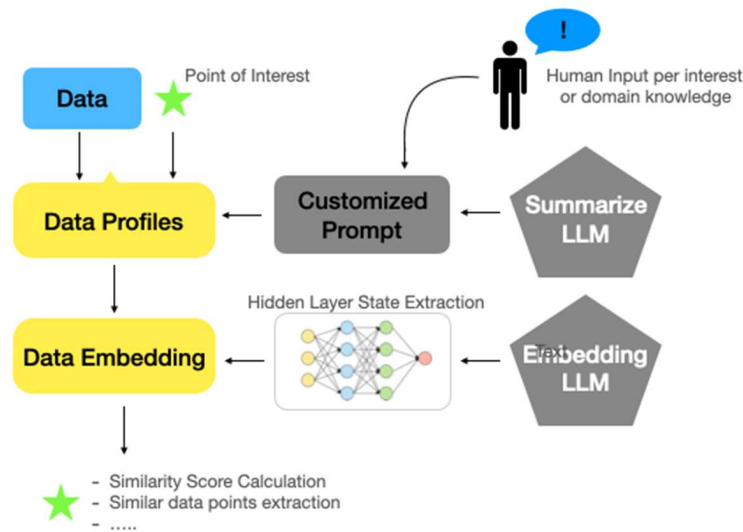

Figure 2. Our proposed Human-in-the-loop Strategy Using Summarization and Hidden State Insights.

**Stage I: Human-in-the-loop Data Summarization**

In the first stage of our methodology, we introduce the Human-in-the-loop Data Summarization process, leveraging the adaptability of Large Language Models (LLMs) to generate data summaries tailored to specific human interests. This stage is pivotal in transforming non-text data into a summarized text format that encapsulates the essence of each data point, significantly reducing complexity and enriching the data with contextual relevance. The core of this stage lies in the interactive customization of the summarization prompts. Users, such as domain experts or stakeholders, input their specific criteria or areas of interest into the system. This input dynamically shapes the LLM's summarization prompts, ensuring that the generated summaries reflect the aspects most relevant to the user's needs. This approach allows for a higher degree of specificity and relevance in the summaries, making subsequent analyses more aligned with user-defined objectives.

To implement this human-centric summarization, we utilize a pre-trained Transformer-based LLM, such as ChatGPT, or a fine-tuned LLM to excel in summarization tasks toward a specific domain. Users can refine their inputs based on preliminary outputs, fostering an iterative loop that hones in on the most informative and relevant data summaries. In order to ensure the effectiveness of this customization, we could employ natural language processing (NLP) techniques to refine and structure user inputs into coherent prompts. This step may involve keyword extraction, semantic analysis, and the generation of query-like structures that are understandable by the LLM. The aim is to bridge the gap between the user's natural language inputs and the model's operational language, ensuring that the prompts are both reflective of the user's intent and optimized for the model's processing capabilities.

After the initial summarization based on the customized prompts, users are presented with the option to review the summaries and refine their inputs. This iterative loop allows for the adjustment of prompts based on the outputs received, enabling a dynamic interaction between the user and the model. Such an approach is particularly beneficial in complex domains where the user's objectives may evolve as new information is uncovered through the summarization process.

The system supports multiple iterations, with each cycle fine-tuning the summarization focus and fidelity. Feedback mechanisms are integrated to capture user satisfaction with the summaries, further informing model adjustments and prompt refinements. This iterative process ensures that the final summaries are highly tailored to the user's specific interests and requirements, making the data more accessible and actionable. The technical architecture supporting this human-in-the-loop process is designed for flexibility and user-friendliness. It includes an intuitive user interface that guides users through the input and refinement process, making it accessible to both technical and non-technical users. Backend components are optimized for rapid processing of custom prompts and generation of summaries, ensuring a smooth and efficient user experience.

In summary, the implementation of Stage I leverages advanced LLM capabilities, interactive customization, and iterative refinement to produce data summaries that are not only concise and information-rich but also meticulously aligned with the user's specific interests and needs. This human-centric approach underscores the potential of LLMs to transform data analysis across various domains, making it a powerful tool for researchers, analysts, and domain experts alike.

**Stage II: Hidden State Extraction**

Following the generation of customized summaries, each summary is processed through a second, advanced LLM specifically tasked with extracting hidden states. Hidden states are essentially vectors that represent the distilled semantic and syntactic essence of the text as understood by the LLM. These vectors capture a level of information and knowledge about the text that is not directly observable in the raw data, encompassing relationships, context, and nuances that are critical for identifying similarity beyond superficial comparisons.

The LLM used for hidden state extraction is selected for its depth and sophistication, typically consisting of numerous layers of Transformer blocks known for their efficiency in capturing linguistic patterns and relationships. At each layer, the model performs complex computations that gradually abstract the text's information into higher-level representations. The specific layers from which hidden states are extracted are carefully chosen based on empirical evidence of their relevance to capturing semantic similarities. This decision is informed by both the literature in the field and experimental validation within the context of our research.

The extracted hidden states serve as a compact, yet information-rich representation of each data point, enabling the effective comparison of seemingly disparate data. To facilitate this comparison, the hidden states are analyzed using techniques such as cosine similarity, which quantifies the likeness between pairs of hidden state vectors. This quantitative measure of similarity allows for the identification of data points that, while different in superficial aspects, share underlying patterns or themes.

Crucially, the process of hidden state extraction is designed to be synergistic with the human-in-the-loop insights from Stage I. The feature-rich representations derived in this stage are inherently reflective of the user-defined focuses and nuances emphasized during the summarization process. This integration ensures that the similarities identified are not only based on the data's inherent characteristics but also aligned with the specific interests and criteria defined by the users, thereby enhancing the relevance and applicability of the findings. In summary, Stage II encapsulates the technical core of our methodology, transforming text summaries into analyzable, feature-rich vectors. This stage bridges the gap between qualitative summaries and quantitative similarity analysis, enabling a sophisticated, nuanced approach to identifying similar data points across diverse datasets.

**Stage III: Similarity Analysis**

Building upon the refined data representations obtained from Stage II, the final part of our methodology involves a meticulous Similarity Analysis. This phase is crucial for operationalizing the insights garnered from the human-in-the-loop summarization and the subsequent extraction of hidden states, ultimately enabling the identification of similar data points across diverse datasets.

The core of the Similarity Analysis lies in harnessing the dense, information-rich vectors—derived from the hidden states—to systematically identify similarities between data points. By employing advanced computational techniques, we transform these abstract representations into actionable insights, facilitating the discovery of patterns and relationships not readily apparent through conventional analysis methods. Our approach utilizes cosine similarity as the primary metric for quantifying the likeness between pairs of data representations. Cosine similarity measures the cosine of the angle between two vectors, providing a scale from -1 to 1 that represents how closely the vectors (and thus, the data points they represent) are related in their semantic and contextual dimensions. This metric is chosen for its effectiveness in high-dimensional spaces, where traditional distance measures may fall short.

To translate similarity scores into actionable classifications of 'similar' or 'not similar,' we establish a threshold value. This threshold is empirically determined based on the capacity or experience of the domain expert, ensuring that it optimally distinguishes between genuinely related and unrelated data points across different domains. The choice of threshold is critical, as it balances the sensitivity and specificity of similarity detection, thereby impacting the utility of the analysis in practical applications. In alignment with the human-in-the-loop ethos of our methodology, domain experts play a crucial role in the final interpretation of similarity analysis results. Their insights are invaluable in contextualizing the quantitative similarity scores within the specific nuances of the domain, enabling a more nuanced understanding and application of the findings. This collaborative approach ensures that the outcomes of the similarity analysis are both technically sound and practically relevant.

The Similarity Analysis phase culminates in a versatile tool for identifying related data points across a variety of contexts, from academic research to industry-specific applications. Whether it's uncovering fraudulent transactions in financial datasets, grouping patients with similar diagnostic images in healthcare, or segmenting customer preferences in marketing data, the methodology offers a scalable, efficient means to unearth hidden similarities grounded in deep semantic and contextual understanding. In essence, the Similarity Analysis not only embodies the analytical capabilities of our method but also highlights its potential to bridge the gap between complex data analysis and real-world decision-making. By meticulously comparing the feature-rich representations of data points, we unlock new possibilities for knowledge discovery and insights across diverse fields, paving the way for innovative applications and enhanced decision-making processes.

## 4. EXPERIMENT

### 4.1. Image Data: MIT Place365 – Scene Understanding Dataset

In this experiment, we leverage the MIT Places365 dataset, a vast and diverse repository of images meticulously categorized into 365 distinct scene types, ranging from natural landscapes to urban environments. Our primary objective is to evaluate the efficacy of our proposed method for identifying similar data points across image datasets using Large Language Models (LLMs). By employing a two-step process that first summarizes the contextual essence of each image before extracting and analyzing the hidden state representations from LLMs, we aim to uncover nuanced similarities among the scenes beyond what conventional pixel-based comparison techniques can achieve. This experimental setup not only tests the versatility and adaptability of our method in handling complex visual data but also aims to contribute to the broader domain of scene understanding by providing a novel approach to categorizing and analyzing vast image datasets. Through rigorous qualitative and quantitative analyses, we anticipate uncovering insights that could significantly enhance automated scene recognition capabilities, ultimately aiding in the development of more intuitive and context-aware computer vision applications.

We first concentrate on kitchen images to explore the potential of Large Language Models (LLMs) in identifying similar data points within a specific scene category. Initially, our methodology involves presenting the model with a kitchen image, then employing our innovative two-step process—combining data summarization and hidden state extraction—to identify and extract images that the LLM recognizes as having similar defining characteristics, primarily ensuring they depict kitchens. Subsequently, we refine our focus towards a more granular feature within these images: the floor type. This phase aims to test the model's ability to discern and match floor types across the extracted kitchen scenes, thereby evaluating the precision of our method in identifying similarities in image datasets based on specific, detailed attributes.

Figure 3 presents the results of an experimental process where images from a scene understanding dataset are summarized using three descriptive tags, with a focus on the functionality of the room depicted. This illustrates the initial phase of the two-step process in which LLMs are employed to generate a condensed, functional profile of image data. From the displayed samples, it is evident that the Large Language Models have successfully identified key functional and design elements within the bathroom category. The tags, such as *#ModernDesign, #SanitaryWare, #shower, and #modern_faucet,* not only highlight the primary function of the room but also reflect specific features that could play a crucial role in the identification of similar images based on functionality and design elements.

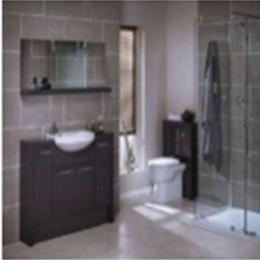

Figure 3. Image profile summarization and similar image extraction for a bathroom scene

These hashtags suggest a standardized approach to tagging, potentially facilitating the retrieval and comparison of images. This could enable the subsequent step of the methodology, wherein these tags serve as a simplified representation for the extraction of hidden states by another LLM. The consistency in identifying the bathroom category across varied design presentations shows promise for the LLM's ability to abstract core functional features from visual data. Moreover, this tag-based profiling may serve as a bridge between the visual characteristics of the scenes and the LLM's text-based processing capabilities, thereby transforming image attributes into a format more amenable to linguistic analysis. This is crucial in enabling LLMs, which are traditionally adept at handling textual data, to process and analyze image data effectively.

Figure 4 showcases the application of a refined image summarization process where each image is tagged with three descriptors concentrating on room functionality and floor color. This experiment extends the previous method by incorporating an aesthetic element — the color of the floor — which adds another layer to the image profiling. In this instance, the Large Language Models (LLMs) were able to recognize and tag not only the room type but also specific design attributes and the color of the floor. Tags such as #beige_floor, alongside #modern, and descriptors like 'Elegant Design' and 'Compact', demonstrate the model's ability to capture both functional and decorative elements of the room. These tags form a more detailed profile of the images that can be used in the subsequent analysis to identify similar scenes.

The tagging process appears to perform well for the bathroom scenes, as indicated by the relevant tags. However, an anomaly is observed in the last image, which is incorrectly tagged as a bedroom despite it being a bathroom scene. This highlights one of the challenges in automated image tagging and summarization: the model's potential to misinterpret scenes when the visual cues are not distinct enough or when the model's training data does not sufficiently cover the variability within the category. In addition, the identification of the floor color as a tagged attribute emphasizes the model's nuanced understanding of image features.

Overall, this experiment underscores the importance of continuously refining the tagging algorithms and the potential need for domain-specific tuning to improve the model's performance across varied datasets and use cases. It also opens up conversations about the depth and breadth of features LLMs should consider when summarizing and categorizing complex image data in the human's interest.

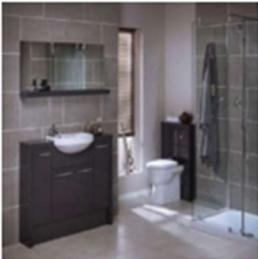
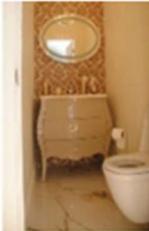
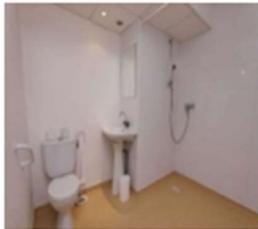
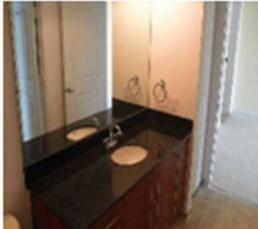

Figure 4. Another image profile summarization, focusing on the floor color, followed by similar images extraction

## 4.2. Tabular Data: AMLSim – Anti-Money Laundering Transaction Dataset

In the second experiment, we explore our application to the domain of anti-money laundering (AML) by utilizing the AMLSim dataset. This dataset provides synthetic but realistic transactional data that simulates typical behaviors observed in money laundering scenarios. Our objective is to construct comprehensive customer profiles from tabular data and subsequently employ another LLM to generate embeddings for the purpose of similarity extraction. To construct customer profiles, we provide the first LLM with structured instructions to capture and summarize essential information from the tabular data. These instructions guide the model to focus on key attributes that characterize customer behavior, such as transaction frequency, amount patterns, beneficiary details, and geographical markers. The model's summarization capability allows us to transform raw transactional data into a textual customer profile that accentuates traits potentially indicative of money laundering activities.

Once the profiles are established, we employ a second LLM trained to generate embeddings from textual data. These embeddings serve as numerical representations of the customer profiles, capturing the nuanced patterns and relationships within the data. By leveraging the LLM's ability to understand and encode contextual information, we aim to produce embeddings that reflect the underlying behavior patterns with a high degree of fidelity. For similarity extraction, we utilize these embeddings to measure the proximity between different customer profiles. Through clustering algorithms or nearest-neighbor searches, we can identify groups of customers exhibiting similar transactional behaviors, which may signify coordinated activities or shared money laundering tactics.

The effectiveness of this approach hinges on the LLMs' ability to abstract and encode complex behaviors from tabular data—a task that poses significant challenges due to the non-sequential nature of such data. In contrast to free-text, where LLMs naturally excel, tabular data requires a structured understanding and an ability to infer relationships between discrete data points. The potential impact of this experiment is significant, offering a novel tool for financial institutions to enhance their AML efforts. By streamlining the profile construction and similarity identification process, LLMs can augment existing systems, allowing for more efficient detection of complex money laundering schemes and contributing to the broader field of financial security.

Figure 5 presents the results of our approach to detecting similar customer activities by converting transactional data into a risk-focused customer profile using Large Language Models (LLMs). Our method first summarizes key transaction attributes into concise, descriptive tags that encapsulate behaviors suggestive of money laundering. The experiment demonstrates the LLM's ability to interpret a complex tabular dataset (customer_df) and distill its contents into three informative tags. These tags—such as *"High frequency cross-border transactions," "Large amounts in different currencies,"* and *"Inconsistent payment formats"*—highlight activities that are commonly associated with money laundering. Notably, the model consistently identifies high-risk features across different examples, suggesting a reliable pattern recognition capability. In the instances marked as "*Is Money Laundering: YES*," the model's tags accurately reflect high-risk factors, indicating that the LLM is successfully identifying patterns associated with money laundering. The tags suggest the LLM's effectiveness in capturing the frequency, international scope, and inconsistency of transactions—traits that are often red flags for financial institutions monitoring illicit activities. However, there is an interesting outcome in the last example where, despite the presence of similar high-risk tags, the model concludes "*Is Money Laundering: NO*." This could indicate a discrepancy in the model's assessment criteria or a possible overfitting to certain data patterns, which requires further investigation.

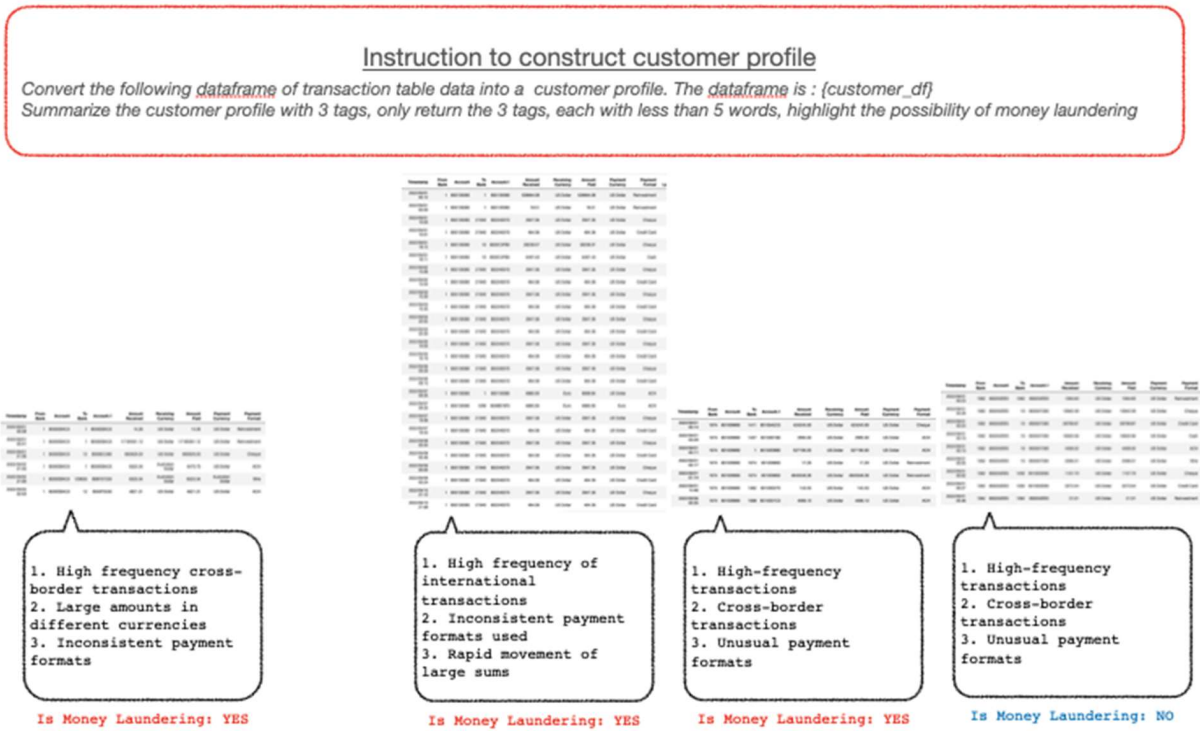

Figure 5. Customer profile summarization from transactional data focusing on anti-money laundering detection

The consistency and accuracy of the tags in the positive examples underscore the model's potential as a tool for financial compliance teams. Yet, the divergence in the last example emphasizes the need for continued refinement of the model and the incorporation of additional data points or contextual information to improve decision-making accuracy. This experiment's insights are significant for the development of LLM-based AML systems. It showcases the power of LLMs to streamline the profiling process in complex, real-world applications, while also highlighting areas for enhancement, such as incorporating cross-reference checks and layering additional analytic methodologies to confirm the LLM's assessments. Future work will involve calibrating the model's sensitivity, validating its conclusions against additional data, and potentially integrating other forms of AI to triangulate findings and increase confidence in the results. The overarching aim is to leverage the LLM's capacity for nuanced data interpretation to support more proactive and granular detection of financial crimes.

## 5. LIMITATION AND FUTURE WORK

The exploration of Large Language Models (LLMs) for identifying similar data points in non-textual domains, while innovative, is not without its constraints. A primary limitation arises from the challenge of model generalization. The LLMs' performance, as evidenced through our experimentation with image and tabular data, might not uniformly extend to disparate types or datasets. Varied domains each bring unique complexities that demand an expansive array of training data to cultivate a well-rounded model comprehension. Another critical concern is the interpretability of LLMs. The opaque nature of these models often leaves users questioning the "how" and "why" behind the model's outputs. This lack of transparency can impede user trust and complicate efforts to diagnose and rectify model errors. Furthermore, the study faced instances of misclassification, signaling a necessity to enhance precision and develop a more nuanced understanding of edge cases. Preprocessing of data, a vital step in our methodology, poses its own challenges. Inaccuracies in these initial stages can have a cascading effect, influencing the overall success of the similarity analysis. Moreover, the computational demands of running LLMs are not insignificant. The resources required to process large datasets may limit the feasibility of this approach for smaller organizations or individual researchers.

In contemplating the future trajectory of our research, several avenues present themselves. An immediate focus would be on increasing the transparency of the LLMs. Introducing techniques to unpack the decision-making process of these models could bolster their interpretability. Expanding the diversity of the datasets in question would likely enhance the robustness of the models, better equipping them to deal with a wider array of data complexities. Refining the models through fine-tuning and transfer learning may offer improvements in both accuracy and operational efficiency. Additionally, the potential of multimodal LLMs that can innately process and understand both visual and textual information holds promise for a more seamless analysis. The integration of human feedback into the model's learning cycle, a 'human-in-the-loop' approach, could serve as a valuable checkpoint for correcting misclassifications and fine-tuning the tagging accuracy. Lastly, to truly scale our approach for practical, widespread use, solutions to mitigate the resource-intensive nature of LLMs must be explored. This could involve seeking out more computationally economical models or leveraging cloud-based infrastructures. Complementing our models with robust validation mechanisms, perhaps through ensemble methods that amalgamate multiple AI systems' insights, could further safeguard against erroneous similarity identifications. In sum, future work will hinge on bridging these gaps, enhancing the methodologies, and ensuring that the foray into using LLMs for data analysis extends its reach, ensuring robust, scalable, and transparent solutions for various real-world applications.

## 6. CONCLUSION

This research presented a pioneering exploration into the use of Large Language Models (LLMs) for the identification of similar data points across non-free text domains, specifically within image and tabular datasets. Our methodology, employing a two-step approach that includes data point summarization and hidden state extraction, has demonstrated significant potential in extracting nuanced similarities that traditional analysis methods might overlook. The experimental findings from the MIT Places365 dataset highlighted the LLM's adeptness at summarizing complex image data into concise tags that represent both the functional aspects and aesthetic features of scenes, specifically within bathroom environments. This capability points to a novel application of LLMs in scene understanding and recognition tasks, extending beyond their conventional textual domain. However, the emergence of outliers, where the model misclassified a room type, signals a need for enhanced model training or supplementary validation techniques. In the realm of tabular data, our experiments with the AMLSim dataset yielded promising results in constructing customer profiles that are indicative of potential money laundering activities. The LLM's proficiency in distilling transactional data into descriptive tags that reflect risk factors pertinent to AML endeavors could revolutionize the way financial institutions monitor and investigate suspicious activities. Notwithstanding these advances, our experiments also surfaced challenges inherent in applying

LLMs across varied data types. In instances of image data, the need to accurately convert visual information into textual summaries that LLMs can process was critical, while for tabular data, ensuring the model's interpretative accuracy remained a pivotal concern. The occasional discordance between the model-generated tags and the expected outcomes suggests avenues for further refinement, such as integrating multimodal data processing or employing ensemble methods to increase the robustness of the models. Overall, our research contributes a novel strategy for leveraging LLMs in data analysis across diverse datasets, offering a scalable and efficient approach for similarity identification. The practical applications of our method have been showcased, providing non-technical domain experts with a tool to swiftly identify data points of interest. Our work opens new pathways for future research in the application of language models beyond the textual domain, advocating for a closer look into multimodal learning and the continued evolution of AI in analytical applications.